\documentclass{sigchi-ext}
\usepackage[T1]{fontenc}
\usepackage{textcomp}
\usepackage[scaled=.92]{helvet} 
\usepackage{graphicx} 
\usepackage{balance}  
\usepackage{booktabs} 
\usepackage{ccicons}  
\usepackage{ragged2e} 

\title{Multiplayer Games for Learning Multirobot Coordination Algorithms}

\numberofauthors{3}
\author{%
  \alignauthor{%
    \textbf{Arash Tavakoli}\\
    \affaddr{Dep. of Computer Science} \\
    \affaddr{University of Southern California} \\
    \affaddr{Los Angeles, CA 90089, USA} \\
    \affaddr{atavakol@usc.edu} } \vfil \alignauthor{%
    \textbf{Haig Nalbandian}\\
    \affaddr{Dep. of Computer Science} \\
    \affaddr{University of Southern California} \\
    \affaddr{Los Angeles, CA 90089, USA} \\
    \affaddr{hnalband@usc.edu} } \vfil \alignauthor{%
    \textbf{Nora Ayanian}\\
    \affaddr{Dep. of Computer Science} \\
    \affaddr{University of Southern California} \\
    \affaddr{Los Angeles, CA 90089, USA} \\
    \affaddr{ayanian@usc.edu} } 
    }

\def\plaintitle{Note Initial
  Caps} \def\plainauthor{First Author, Second Author, Third Author,
  Fourth Author, Fifth Author, Sixth Author}

\def\plainkeywords{
  Multirobot systems; distributed robotics; networked games; games for a purpose; online multiplayer games; social computation; crowdsourced coordination; human computation; human-inspired robotics.}

\hypersetup{%
  pdftitle={\plaintitle}, pdfauthor={\plainauthor},
  pdfkeywords={\plainkeywords}, }

\begin{document}

\maketitle

\RaggedRight{} 

\begin{abstract}
  Humans have an impressive ability to solve complex coordination problems in a fully distributed manner. This ability, if learned as a set of distributed multirobot coordination strategies, can enable programming large groups of robots to collaborate towards complex coordination objectives in a way similar to humans. Such strategies would offer robustness, adaptability, fault-tolerance, and, importantly, distributed decision-making. To that end, we have designed a networked gaming platform to investigate human group behavior, specifically in solving complex collaborative coordinated tasks. Through this platform, we are able to limit the communication, sensing, and actuation capabilities provided to the players. With the aim of learning coordination algorithms for robots in mind, we define these capabilities to mimic those of a simple ground robot.         
  
\end{abstract}

\vspace{0.8cm}




\section{Introduction}
Large teams of humans demonstrate an exceptional ability to solve collective problems in a distributed manner. Yet distributed algorithms for coordinating large teams of robots to work toward a collective objective are challenging to develop. A machine learning approach could be useful in developing such algorithms, but would require a large amount of data on the high-level decision-making process in human groups. It would also require data specialized so that it could be used on a multirobot platform. 

\begin{marginfigure}[-54pc]
  \begin{minipage}{\marginparwidth}
    \centering
    \includegraphics[width=0.9\marginparwidth]{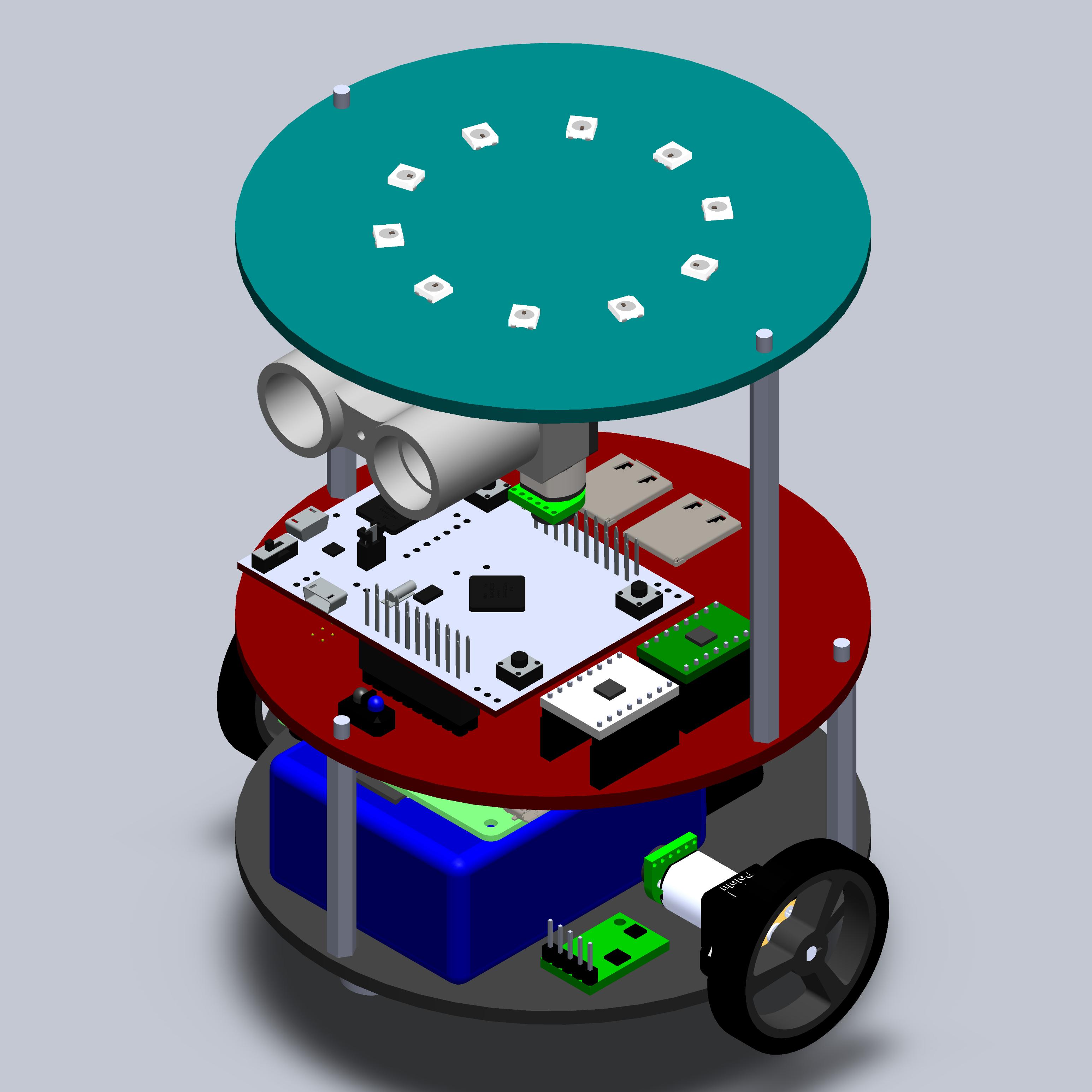}
    \caption{Our in-house developed differential-drive robot platform whose capabilities were considered for the design of our online multiplayer games.}~\label{fig:actbot}
  \end{minipage}
\end{marginfigure}

One way to collect such data is by using crowdsourced multiplayer games. Crowdsourced games have demonstrated great potential as a strategy for data-collection to solve a variety of complex computational problems while leveraging human abilities \cite{chernova2010crowdsourcing, cooper2010predicting, kearns2012experiments, shahrokhi2015stochastic, von2009human, von2004labeling}. Besides providing a method for collecting large amounts of data, control over the game interface enables limiting sensing, communication, and actuation capabilities provided to players, so they mimic those of a robot. This is important since we are interested in developing algorithms that would be implemented on simple ground robots (e.g. our platform in Fig.~\ref{fig:actbot}) without requiring additional sensory or communication infrastructure. In addition, using a multiplayer gaming platform enables us to investigate human decision-making in group tasks over a range of scenarios with minimal modification efforts to the platform.  

In this work, we describe the current (second) version of our online multiplayer gaming platform, which was used in a single investigation with 25 simultaneous players (pilot 2). This version has evolved based on experiences and results from a prior version used in a single investigation (pilot 1) with 15 simultaneous players. 
We describe the platform and discuss some fundamental design considerations. Furthermore, we show some representative results from pilot 2.




\begin{marginfigure}[-20pc]
  \begin{minipage}{\marginparwidth}
    \centering
    \includegraphics[width=0.9\marginparwidth]{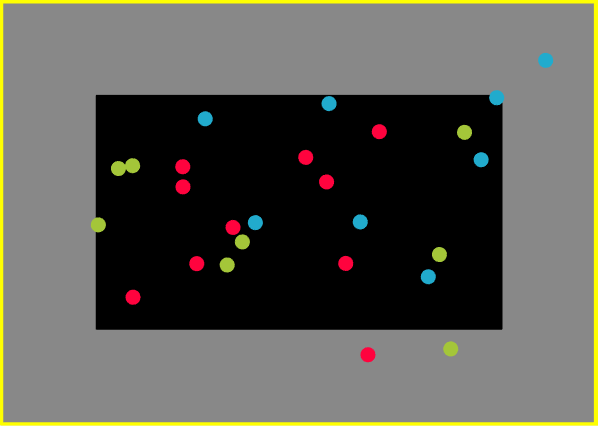}
    \caption{A top-view snapshot of the arena with 26 players. The yellow borders are the arena's walls, black rectangle is the region of the arena visible in the field of view of the Overhead View, and the grey margin indicates the parts of the arena which are not visible in the Overhead View.          
      }~\label{fig:topwide}
  \end{minipage}
\end{marginfigure}

\section{Purpose of the Game}
The purpose of our game is to collect data for analysis of human group behaviors, and learn algorithms and strategies from that data to solve challenging multirobot coordination problems. In our game, we have targeted the distributed multirobot formation problem: \emph{how can we program a group of robots to form a global pattern without receiving commands from a central station and without explicit communication?} This is a challenging research problem in the field of multirobot systems. The goal of each instance of play was the collective formation of a pattern. 

Since the intention of the game is to collect data for our research, the game scenarios and setup are designed to fit our research problem, which is to find distributed algorithms for coordinating a team of robots toward a formation objective, with no explicit communication. Furthermore, the information made available to players in the game is designed to mimic data available to the physical platform we plan to use for practical application of the learned robot behaviors: signals from a proximity sensor and an overhead image that is broadcasted to them at a fixed frequency.

\section{Gameplay and Interface Design Considerations}
The gaming platform was designed so the sensing and communication capabilities of the human-controlled agents would be similar to those of our in-house robotic platform (Fig.~\ref{fig:actbot}). Additionally, as mentioned earlier, the game's scenarios should match our multirobot coordination problem. Thus, interface design and gameplay are tightly coupled. 

\begin{figure}[t]
  \includegraphics[width=1\columnwidth]{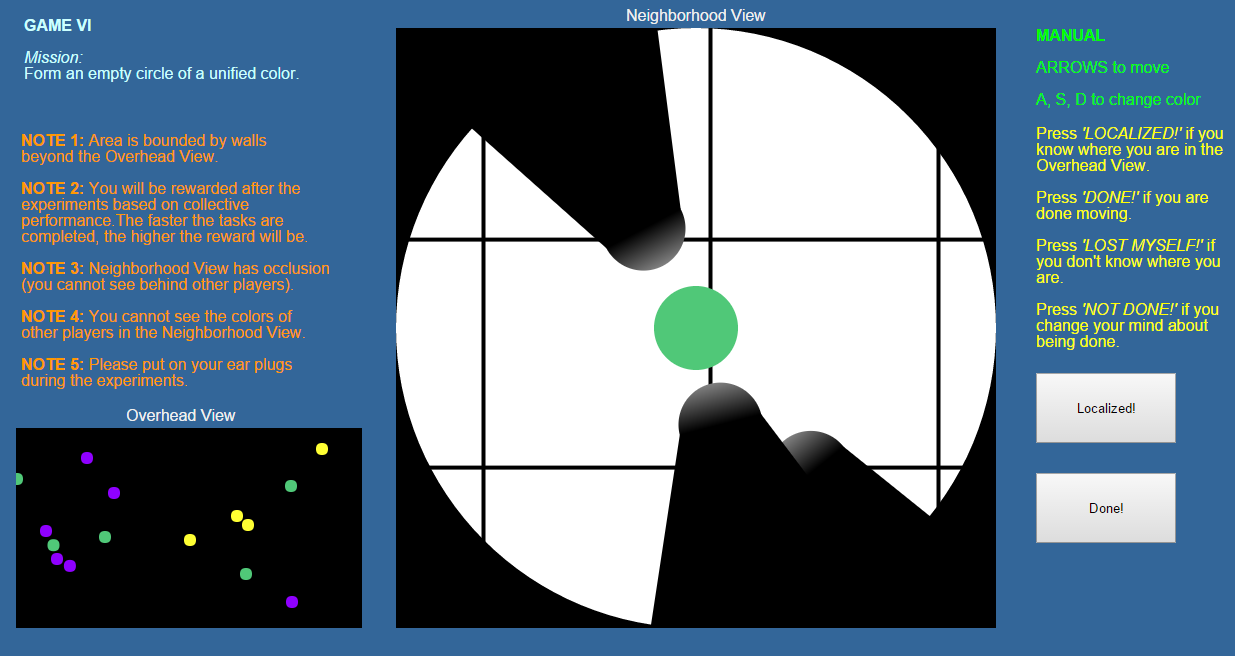}
  \caption{Sample screenshot of the player's GUI. Game instance includes both Neighborhood and Overhead Views.}~\label{fig:playerGUI}
\end{figure}

In each session, the task of the multiple simultaneously playing participants is to collectively form into a specified pattern.  In our research problem, we are investigating how humans use different abilities (communication, sensing, etc.) in order to accomplish the collective task. 
Thus, the game is played multiple times (referred to as instances) by participants in one experimental session. In order to investigate how players use their capabilities, capabilities are altered in each instance to investigate the ability of the same participants to complete the formation missions in the absence of one or a combination of their capabilities. The following gaming agent capabilities were implemented: 
\begin{itemize}\compresslist%
\item \textbf{\textit{Local Sensing.}} Agents have fixed-range omnidirectional local perception with occlusions (\textit{Neighborhood View}), similar to the robot's 360-degree rotating laser rangefinder proximity sensor (see Fig.~\ref{fig:playerGUI}).
\item \textbf{\textit{Global Sensing.}} Agents receive global information (\textit{Overhead View}) that imitates a WiFi-broadcasted image from an overhead camera in the practical application (see Fig.~\ref{fig:playerGUI}). The Overhead View image updates at 1 Hz (0.2 Hz in pilot 1), to limit the use of global information, which can be costly in multirobot systems. 
\item \textbf{\textit{Inter-Agent Communication.}} Explicit communication was not provided.
\item \textbf{\textit{Color-Switching.}} Agents can switch their color; this can be seen in the agent's own Neighborhood View, and by other agents in the Overhead View. 
\item \textbf{\textit{Motion}} Agents can move in the eight compass directions at a fixed velocity of 18 pixels/s, corresponding to a simple ground robot's average velocity of approximately 0.2 m/s. To create a better sense of motion when an agent is moving, a patterned floor was added only to the Neighborhood View.
\end{itemize}
%
%
%
%
The agents' environment is a collective rectangular workspace (called the \textit{arena}) bounded by walls in which agents could move freely using the controls available to them (see Table~\ref{tab:controls}). The arena is partially observable to players through the Overhead View (see Fig.~\ref{fig:topwide}). 

\begin{margintable}[-40pc]
  \begin{minipage}{\marginparwidth}
    \centering
    \begin{tabular}{r r}
      & {\small \textbf{Control Keys}} \\
      \toprule
      Motion & Arrow Keys \\ 
      \midrule
      Color Change & A,S,D \\
      \bottomrule
    \end{tabular}
    \caption{Keys for controlling the agent.}~\label{tab:controls}
  \end{minipage}
\end{margintable}


\section{Technological Aspects}
We designed the multiplayer gaming platform as a web application comprised of three major components: (1) the \textit{admin GUI} (Graphical User Interface), which is used by the game's administrator to initiate a new instance of the game; (2) the \textit{player's GUI} (see Fig.~\ref{fig:playerGUI}), which players use to participate in the game; and (3) the \textit{server}, a \textit{Node.js} application. The server sets the players' agents in a common arena and updates the global state for all agents in the game, as well as records timestamped decision trajectories (i.e., position and color) for each player at 10 Hz. 

The player's GUI was designed minimalistically to ensure the platform is easy to use. The player's GUI, based on each game instance's task specification, provides agents with the appropriate capabilities, including: the Overhead View, the Neighborhood View, color-switching, and motion control primitives.     
The interface was written in \textit{HTML5}, utilizing the \textit{canvas} element, and \textit{JavaScript}, to draw and update the agents' views. 

For our research, it is essential to track each player's data in a single session across multiple instances of gameplay, and label it with a unique identifier. To this end, \textit{browser cookies} are used to store a unique identifier on each player's browser that allows us to store session data (across all instances) for each player.

\section{Results}
\begin{marginfigure}[-12pc]
  \begin{minipage}{\marginparwidth}
    \centering
    \includegraphics[width=0.9\marginparwidth]{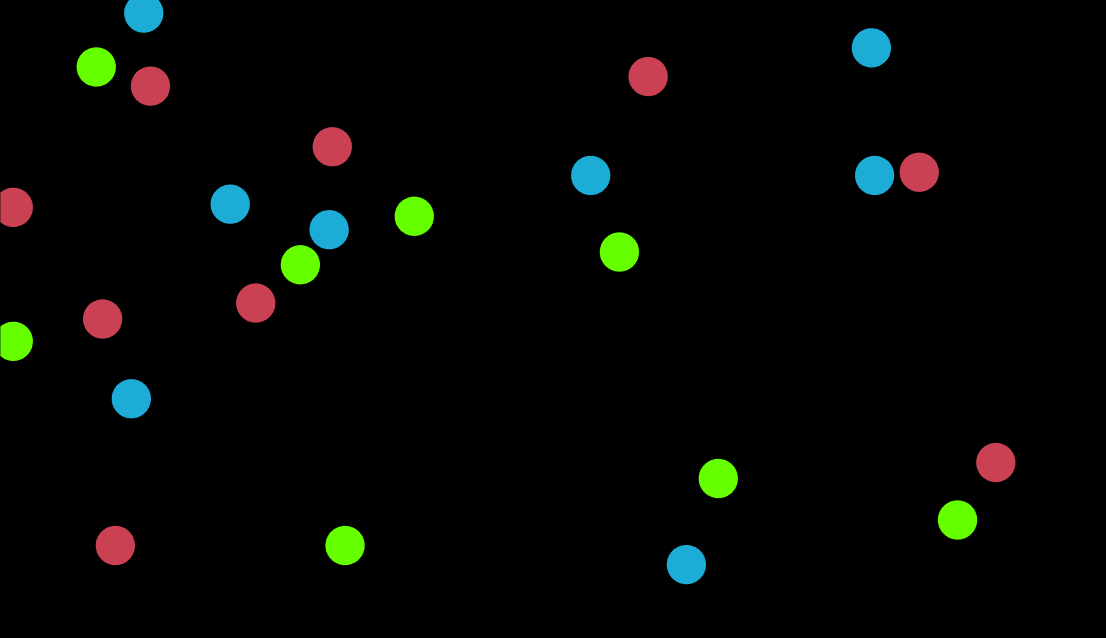}
    \label{5i}
    \vspace{2pt}
    \includegraphics[width=0.9\marginparwidth]{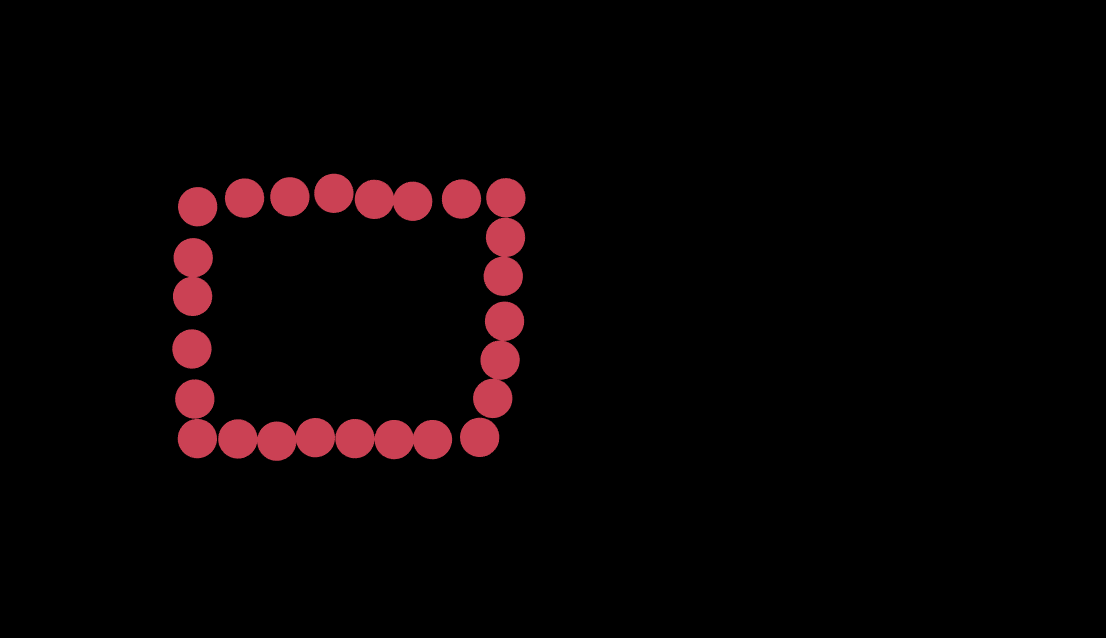}
    \label{5f}
    \caption{Initial random configuration (Top) and the completed formation (Bottom) for one instance of the game in pilot 2.}\vspace{-5pt}
    \label{fig:results}
  \end{minipage}
\end{marginfigure}

Figure \ref{fig:results} shows the initial state and the final state just after the objective was fulfilled in a representative instance that we ran as part of our pilot 2 investigation \cite{Tavakoli:2016}. Here the objective of the game was for the team to organize into a rectangle of unified color using both the Overhead and Neighborhood Views. Note that in this case, color consensus must also be reached.    

In instances with only the Overhead View available, duration of the task was much lower than when only the Neighborhood View was available. 
In an instance with Neighborhood View only, players were able to form a rectangle, even though the local view only showed a small portion of that rectangle (colors were fixed in this instance). This task took 513s, whereas the rectangle took only 62s with both Overhead and Neighborhood Views as well as color switching. With  Overhead View only and color switching, the task took 149s,
indicating a strong effect of global feedback. 

While players were not given the ability to communicate explicitly with others, they developed novel signaling techniques that can easily be used on a robotic system. For example, players got the attention of other agents by changing colors rapidly or repeatedly bumping into other agents. 



\section{Conclusions}
We described a multiplayer gaming platform that we have developed to investigate human group behaviors. These investigations have helped us gain a better understanding of human coordination in collective distributed formation tasks. We intend to further develop and improve the platform for a wider range of tasks and  to target larger numbers of participants through established online crowdsourcing platforms.

\section{Acknowledgements}
This work was partly supported by NSF CAREER IIS-1553726. Arash Tavakoli acknowledges support from the Viterbi Graduate School Ph.D. Fellowship.

\balance{} 

\bibliographystyle{SIGCHI-Reference-Format}
\bibliography{sample}

\end{document}